\documentclass[conference]{IEEEtran}
\IEEEoverridecommandlockouts
\usepackage{cite}
\usepackage{amsmath,amssymb,amsfonts}

\usepackage{graphicx}
\usepackage{textcomp}
\usepackage{xcolor}
\usepackage{multirow}
\usepackage{amsmath}
\usepackage[utf8]{inputenc} % allow utf-8 input
\usepackage[T1]{fontenc}    % use 8-bit T1 fonts
\usepackage{hyperref}       % hyperlinks
\usepackage{url}            % simple URL typesetting
\usepackage{booktabs}       % professional-quality tables
\usepackage{amsfonts}       % blackboard math symbols
\usepackage{nicefrac}       % compact symbols for 1/2, etc.
\usepackage{microtype}      % microtypography
\usepackage{xcolor}         % colors
\usepackage{colortbl}
\usepackage{graphicx}
\usepackage{amsmath}
\usepackage{algorithm}
\usepackage{multirow}
\usepackage{algpseudocode}
\usepackage{adjustbox}
\usepackage{booktabs}
\usepackage{multirow}
\definecolor{headergray}{gray}{0.85}
\definecolor{rowgray}{gray}{0.95}
\definecolor{bestgreen}{rgb}{0.0, 0.6, 0.0}
\def\BibTeX{{\rm B\kern-.05em{\sc i\kern-.025em b}\kern-.08em
    T\kern-.1667em\lower.7ex\hbox{E}\kern-.125emX}}
\begin{document}

\title{CMAT: A Multi-Agent Collaboration Tuning Framework for Enhancing Small Language Models}

\author{
% 第一行作者（前三个添加星号）
\IEEEauthorblockN{
Xuechen Liang\textsuperscript{1,*},
Yangfan He\textsuperscript{2,*},
Meiling Tao\textsuperscript{3,*},
Yinghui Xia\textsuperscript{4}
}

% 第二行作者
\IEEEauthorblockN{
Jianhui Wang\textsuperscript{5},
Tianyu Shi\textsuperscript{6},
Jun Wang\textsuperscript{7},
Jingsong Yang\textsuperscript{8}
}

% 机构信息区块
\IEEEauthorblockA{
\textsuperscript{1}East China Jiaotong University, China \hfill
\textsuperscript{2}University of Minnesota - Twin Cities, United States \\
Email: lxc974464657@outlook.com \hfill
Email: he000577@umn.edu
}

\IEEEauthorblockA{
\textsuperscript{3}Guangdong University of Technology, China \hfill
\textsuperscript{4}Autoagents.ai, China \\
Email: 3221010067@mail2.gdut.edu.cn \hfill
Email: vtx@autoagents.ai
}

\IEEEauthorblockA{
\textsuperscript{5}University of Electronic Science and Technology of China, China \hfill
\textsuperscript{6}University of Toronto, Canada \\
Email: 2022091605023@std.uestc.edu.cn \hfill
Email: tianyu.sh19@mail.mcgill.ca
}

\IEEEauthorblockA{
\textsuperscript{7}East China Normal University, China \hfill
\textsuperscript{8}Autoagents.ai, China \\
Email: wongjun@gmail.com \hfill
Email: edward.yang@autoagents.ai
}

% 共同一作脚注
\IEEEauthorblockA{
\raisebox{0pt}[0pt][0pt]{\rule{0pt}{10pt}} % 垂直间距调整
\textsuperscript{*} Equal contribution.
}
}
\maketitle

\begin{abstract}
Open large language models (LLMs) have significantly advanced the field of natural language processing, showcasing impressive performance across various tasks. Despite the significant advancements in LLMs, their effective operation still relies heavily on human input to accurately guide the dialogue flow, with agent tuning being a crucial optimization technique that involves human adjustments to the model for better response to such guidance.
Addressing this dependency, our work introduces the TinyAgent model, trained on a meticulously curated high-quality dataset. We also present the Collaborative Multi-Agent Tuning (CMAT) framework, an innovative system designed to augment language agent capabilities through adaptive weight updates based on environmental feedback. This framework fosters collaborative learning and real-time adaptation among multiple intelligent agents, enhancing their context-awareness and long-term memory. In this research, we propose a new communication agent framework that integrates multi-agent systems with environmental feedback mechanisms, offering a scalable method to explore cooperative behaviors. Notably, our TinyAgent-7B model exhibits performance on par with GPT-3.5, despite having fewer parameters, signifying a substantial improvement in the efficiency and effectiveness of LLMs. 
\end{abstract}


\section{Introduction}

In the rapid development of the field of artificial intelligence, large language models (LLMs) such as BERT and GPT-4 ~\cite{openai:gpt4} have become important cornerstones of natural language processing (NLP). These models utilize the Transformer architecture and effectively capture long-distance dependencies through multi-head self-attention mechanisms, demonstrating strong capabilities across various NLP tasks. With technological advancements, the performance and application scope of LLMs continue to expand, promising significant improvements in computational efficiency and functionality, including anticipated advanced features such as self-improvement, self-checking, and sparse expert models~\cite{liu2023agentbench}. 

\begin{figure}[t!]
    \centering
    \includegraphics[width=\linewidth]{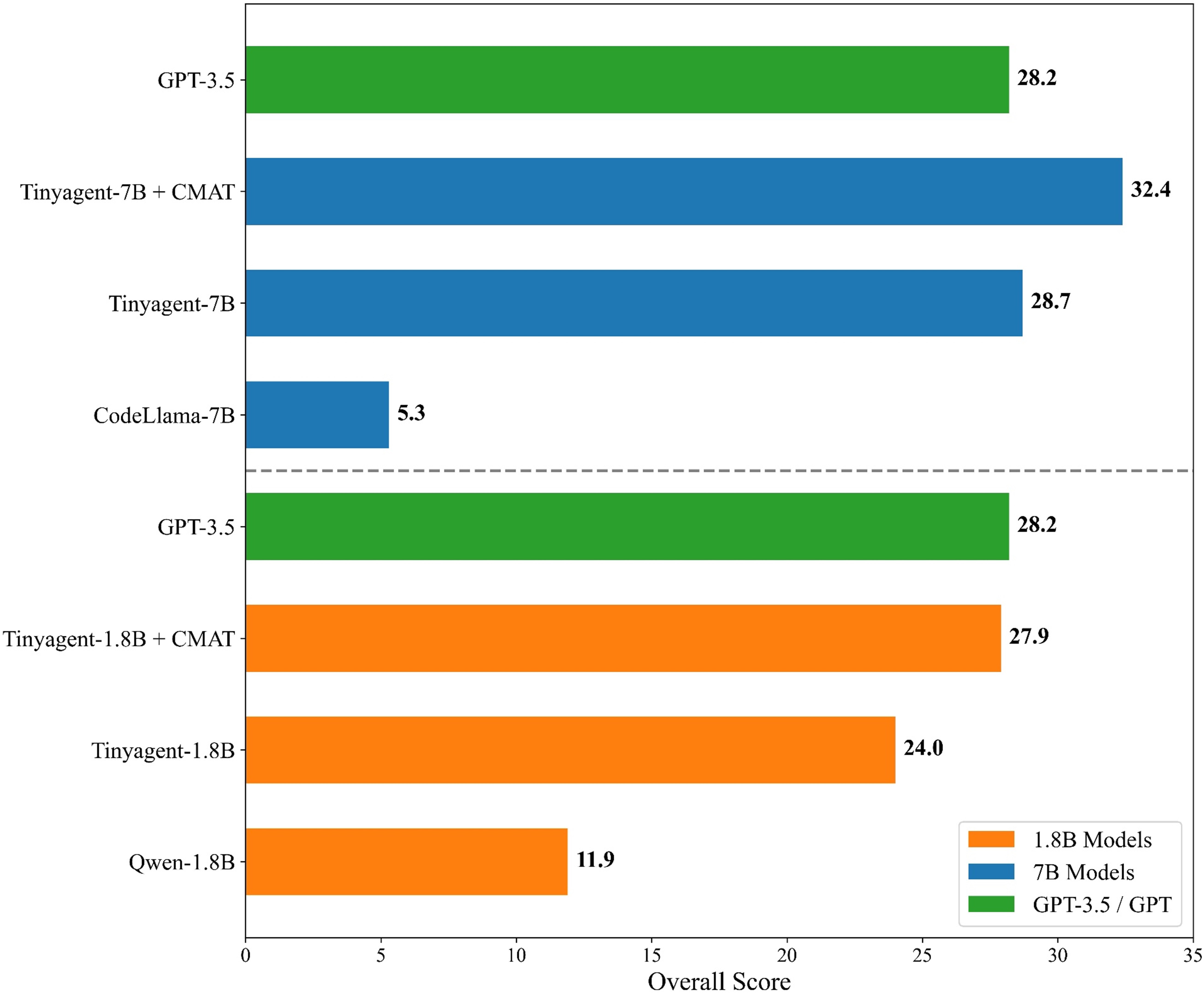}
    \caption{
TinyAgent demonstrates outstanding performance, comparable to that of GPT-3.5. TinyAgent is a series of models fine-tuned based on Qwen~\cite{bai2023qwen} and Codellama~\cite{roziere2023code}.}
    \label{fig:example3}
    \vspace{-20pt}
\end{figure}

However, it is noteworthy that the success of these models largely depends on human input to guide the correct dialogue. This dependency requires users to provide relevant and precise prompts based on their intentions and the feedback from the chat agent, raising a critical question: 
%Can we replace human intervention with autonomous communication agents capable of steering conversations towards task completion with minimal human supervision? 
\textit{\textbf{Can we replace human intervention with autonomous communication agents capable of steering conversations towards task completion with minimal human supervision?}}
%With the swift progression of artificial intelligence, large language models (LLMs) such as BERT~\cite{devlin2019bert} and GPT-3.5 have emerged as pivotal in the realm of natural language processing. Trained on extensive datasets, these models are adept at comprehending and generating human language, showcasing remarkable proficiency across a spectrum of tasks~\cite{Zhou2023A}. 
%Yet, deploying these models in real-world settings encounters hurdles, including constraints on computational resources, data biases, and a lack of model robustness~\cite{Goetze2021Bigger}.
%To surmount these obstacles, the concept of multi-agent systems (MAS) has been introduced. These systems, by promoting cooperation and interaction between agents, not only bolster processing efficiency but also amplify the adaptability and flexibility of the system~\cite{Jang2021Towards,Mokander2023Auditing}.Though they have performed well in traditional NLP tasks and largely advanced the development of LLMs. The performance gap in agent tasks hampers the advancement of in-depth LLM research and community innovation.

Our research is driven by the need to overcome the significant challenges faced by LLMs in real-world deployments, particularly the high computational resource requirements, data biases, and lack of robustness. These issues limit their applicability in resource-constrained environments and highlight the urgency of enhancing model efficiency and adaptability~\cite{abid2021large,du2022shortcut}. As demonstrated by Figure~\ref{fig:example3}, we aim to address these limitations by optimizing models and training methods to enable smaller models to match the performance levels of larger models. Additionally, recognizing the potential of MAS to improve processing efficiency and system adaptability through agent cooperation, we seek to develop a collaborative agent framework. This framework aims to facilitate effective cooperation among agents, thereby overcoming the performance gap and propelling further research and innovation in the field of LLMs~\cite{Ferry2018CloudMF:,Talwar2005Comparison}. In our experiments, we evaluated the capabilities of large models with and without the use of prompts and observed that low-quality prompts can significantly degrade model performance. Consequently, we propose the Collaborative Multi-Agent Tuning (CMAT) framework.

\begin{figure*}[t]
    \centering
    \includegraphics[width=0.9\textwidth]{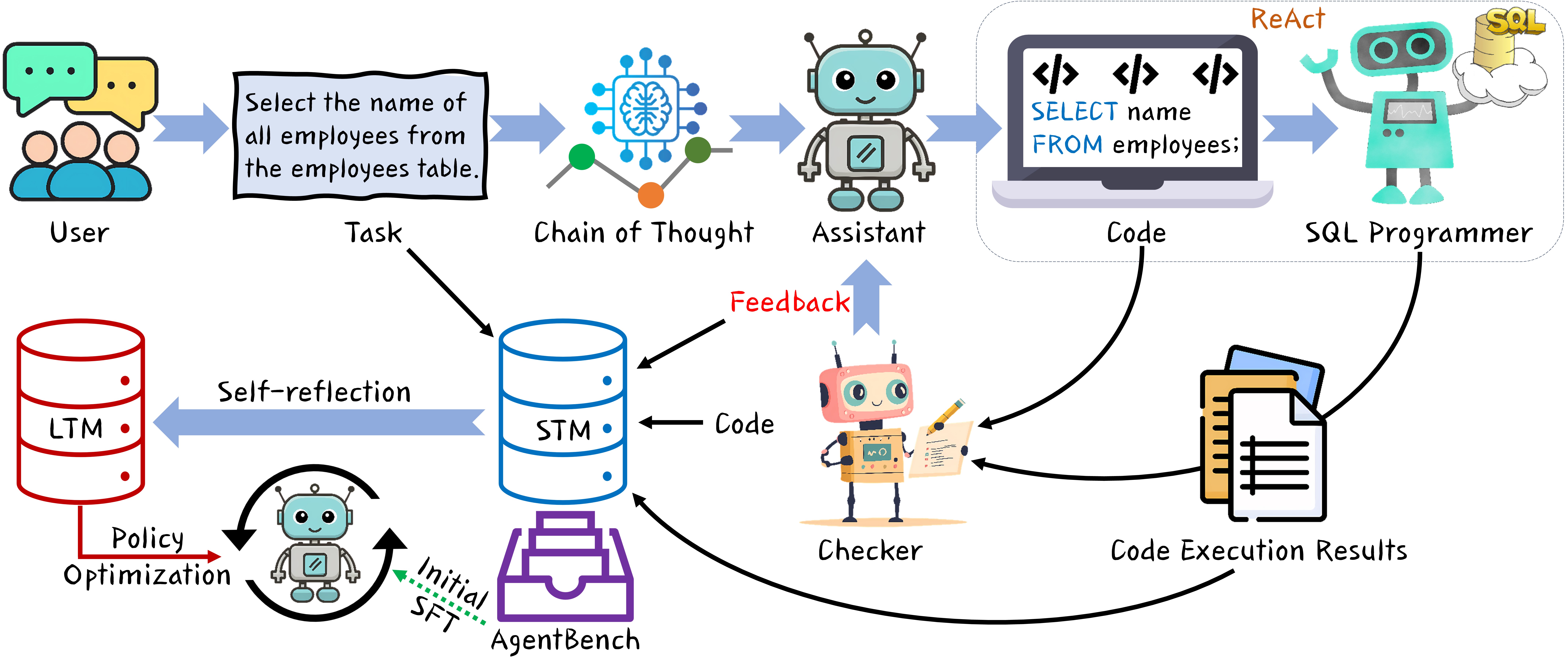}
    \caption{In the CMAT framework, the user assigns tasks to an assistant, which generates SQL commands based on short-term and long-term memories: short-term memory provides immediate context from trajectory history, while self-reflective outputs are stored as long-term memory. The checker verifies the correctness of SQL commands before they are executed in the environment.}
    \label{fig:example1}
    \vspace{-15pt}
\end{figure*}

%In our approach, we introduce a  meticulously curated dataset. This dataset serves as the foundation for training our series of TinyAgent models, which are custom-designed to operate efficiently within the constraints of limited computational resources. To further enhance the effectiveness of these models in practical applications, we propose the Collaborative Multi-Agent Training (CMAT) framework. The establishment of the CMAT framework is predicated on the principle of fostering collaborative learning among multiple TinyAgents. This approach enables the leveraging of each TinyAgent's unique strengths while mitigating their individual weaknesses through collective effort. Such collaboration not only elevates the overall performance of the system but also addresses the adaptability and robustness challenges that are often encountered with singular, larger models. By integrating the CMAT framework with our TinyAgent models, we aim to achieve a paradigm shift towards more resource-efficient, adaptable, and resilient solutions in natural language processing.

% Our research is primarily focused on addressing the challenges associated with the reliance on human input in the deployment of Large Language Models (LLMs), which often limits their robustness and adaptability in dynamic real-world scenarios. To enhance the autonomy and efficiency of LLMs, we are developing the Collaborative Multi-Agent Training (CMAT) framework. This innovative approach aims to reduce dependency on human-generated data by enabling a system of multiple agents to learn and improve collaboratively.

The CMAT framework introduces a structured environment where individual agents, each with specialized roles and capabilities, work together to process information, make decisions, and solve complex tasks~\cite{hernandez2017new}. By sharing insights and learning from interactions within this multi-agent ecosystem, the framework allows for a more scalable and flexible approach to training LLMs~\cite{lewis2017deal}. This collaborative effort not only helps in bridging the gap in performance between smaller and larger models but also fosters a more resilient system capable of adapting to new challenges without extensive human intervention~\cite{kaplan2020scaling}. Through CMAT, we aim to push the boundaries of what is possible with LLMs, making them more accessible and effective for a wider range of applications~\cite{rajpurkar2018know}.

The main contributions of our work are as follows:
\begin{itemize}
    \item We propose the CMAT framework which represents an innovative approach that allows for dynamic and real-time memory updates within multi-agent systems.
    \item We design a novel role-playing mechanism for precise task allocation and enhanced agent communication, significantly boosting overall performance and cooperation.
    \item We evaluated the fine-tuned TinyAgent models across multiple agent tasks, finding that in certain scenarios, their performance rivals that of advanced LLMs like GPT-4 and agentlm ~\cite{zeng2023agenttuning}, demonstrating the potential efficiency and capabilities of compact models.
\end{itemize}

% \begin{algorithm}

% \caption{CMAT Framework}
% \begin{algorithmic}[1]

% \State \textbf{Init:} LLMs, $U$, $A$, $C$, $T$, $\pi_0$, $\text{mem}$.

% \For{$t \in T$}
%   \State \textbf{Exec:} Assign $A$ and $C$.
%   \While{$\neg\text{complete}(t)$}
%     \State $a \gets A \leftrightarrow \text{LLMs}; \text{exec}(a) \to (s', r)$.
%     \If{$C \text{ verifies } a$}
%       \State $\text{update}(s', \text{mem} \cup \{(s', r)\})$.
%     \Else
%       \State $\text{adjust}(\pi, \text{LLMs}); \text{retry}(t)$.
%     \EndIf
%   \EndWhile
%   \State \textbf{Update $\pi$:} With $C$ feedback.
% \EndFor

% \State \textbf{Complete:} Check all $T$; \\
% \textbf{Output:} $\{(s, \text{eval})\}_{t \in T}$.

% \end{algorithmic}
% \end{algorithm}
\section{Related Work}

\subsection{LLMs Applications in a Multi-Agent Framework}
 We explore the applications of LLMs within multi-agent systems, highlighting their role versatility as users, assistants, and checkers, and their capability to offer bespoke support and solutions across such environments~\cite{de2023emergent,talebirad2023multi}. LLMs showcase remarkable adaptability to tasks through methods like supervised fine-tuning and real-time feedback learning, notably in tasks that require a sophisticated understanding and execution related to operating systems or databases~\cite{christianos2023pangu,li2023camel}. Furthermore, LLMs are adept at enhancing communication and collaboration among agents, a critical component for addressing complex issues that necessitate multi-role coordination~\cite{zhao2021adaptive}.
 Nevertheless, LLMs encounter specific challenges within multi-agent frameworks, especially in situations that demand a nuanced contextual comprehension and sustained memory retention, as well as adapting to fast-evolving environments and unforeseeable tasks~\cite{Diallo2020Coordinated}.
 Issues such as data bias, security concerns, and the intricacies of crafting effective protocols for multi-agent cooperation stand as significant hurdles in this domain~\cite{Zhang2017A,García2015Periodic}.
 Thus, by summarizing LLMs' roles in multi-agent frameworks, we underscore the critical need for continued innovation and research exploration, aimed at overcoming these technological hurdles and leveraging the full potential of LLMs in complex systems~\cite{lu2020blockchain}.
% To enhance LLMs in multi-agent systems, we've implemented memory modes, including long-term support and short-term memory with environmental feedback. This allows LLMs to better interact, learn, and adapt in dynamic environments, leveraging past experiences and responding to changes swiftly. Memory modes boost LLMs' adaptability, efficiency, and collaborative potential in complex systems.

To enhance the adaptability and collaborative capabilities of LLMs in multi-agent systems, we've implemented memory modes, including long-term support and short-term memory with environmental feedback~\cite{Liang2016Multiagent}. This allows LLMs to better interact, learn, and adapt in dynamic environments, leveraging past experiences and responding to changes swiftly.

 %To further enhance the application of LLMs in multi-agent systems, we have introduced the concept of memory modes, including long-term memory support for assistants and short-term memory and environmental communication feedback mechanisms. This added functionality enables LLMs to interact and learn more effectively within multi-agent environments, maintaining a record of past interactions and solutions through long-term memory, while adapting to rapidly changing environments and tasks using short-term memory and instant feedback. The introduction of memory modes not only improves the adaptability and efficiency of LLMs in handling complex tasks but also opens new pathways for communication and collaboration within multi-agent systems, further advancing the comprehensive potential of utilizing LLMs in complex systems.

\begin{figure*}[t]
    \centering
    \includegraphics[width=0.85\textwidth]{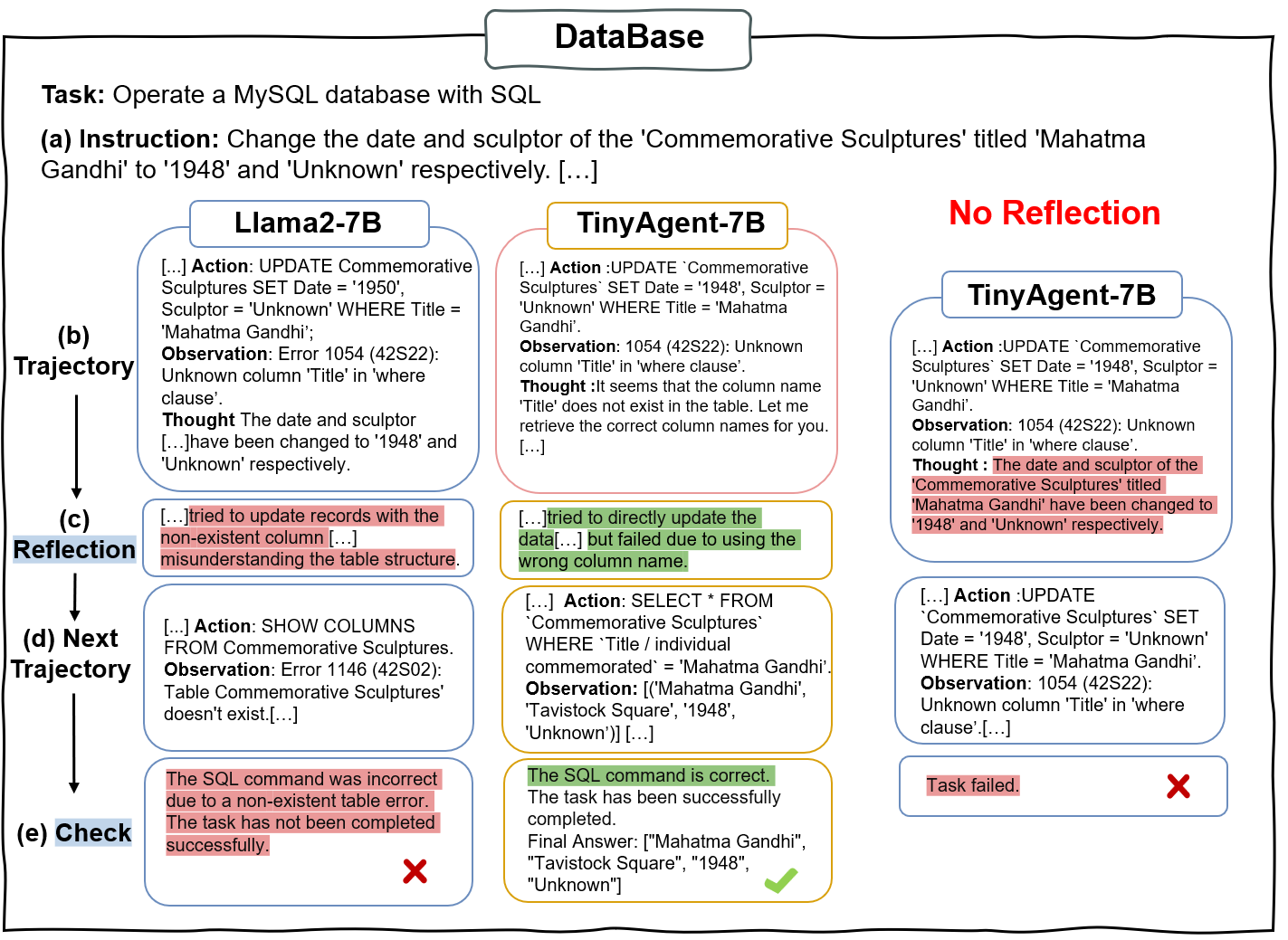}
    \caption{Comparative study of Llama-2-7b and TinyAgent-7b in DataBase cases. (1) In DataBase tasks with a reflection mechanism, Llama-2-7b still made errors after reflection, while TinyAgent-7b adjusted its operations after reflecting on its first failed attempt. (2) Without a reflection mechanism, TinyAgent-7b repeated the same operation and ultimately failed to complete the task.}
    \label{fig:example2}
    \vspace{-15pt}
\end{figure*}

\subsection{The tuning method for LLMs}

 The main tuning methods include supervised fine-tuning and reinforcement learning~\cite{Ouyang2022Training}. Supervised fine-tuning enhances performance by training models on specific task datasets, and is especially suitable for tasks such as natural language understanding (NLU)~\cite{Howard2018Universal}. On the other hand, reinforcement learning, guided by reward mechanisms, is suitable for handling complex and variable tasks~\cite{Mnih2015Human-level}.
The effective combination of these two methods can significantly improve the performance of LLMs in various tasks. Notably, LLMs of reduced scale, such as those encompassing 1.8 billion parameters, can achieve performance levels akin to those of models with greater parameter counts, like 6 billion parameters, when supported by high-quality datasets~\cite{Stiennon2020Learning}. This demonstrates that excellent data quality and appropriate tuning strategies play a decisive role in the performance of LLMs. Therefore, investing efforts in improving data quality and choosing the right tuning methods is essential for achieving optimal performance of LLMs in various application scenarios~\cite{Howard2018Universal}. Through our work combining supervised fine-tuning with reinforcement learning, we've notably advanced LLM performance across a spectrum of tasks, showcasing significant improvements in task-specific benchmarks~\cite{Ouyang2022Training}.

\section{Methodology}

The Collaborative Multi-Agent Language Model Tuning (CMAT) framework improves decision-making, controllability, and efficiency in complex systems by coordinating three roles: User ($\mathcal{U}$), Assistant ($\mathcal{A}$), and Checker ($\mathcal{C}$). We fine-tune language models using LoRA~\cite{hu2021lora}, P-Tuning~\cite{lester2021power}, and integrate ideas from RLHF~\cite{vazquez2019reinforcement}, Chain of Thought (CoT), and ReAct to enhance reasoning and adaptability.

\subsection{Agent Roles and Actor-Critic Dynamics}

In CMAT, the User ($\mathcal{U}$) provides inputs, the Assistant ($\mathcal{A}$) acts as the Actor generating actions, and the Checker ($\mathcal{C}$) serves as the Critic providing feedback. At time $t$, given input $\mathbf{x}_t$, the Assistant produces action $\mathbf{a}_t$ based on its policy $\pi_{\theta_{\text{actor}}}$. The Checker evaluates this action and returns feedback $f_t$, guiding continuous policy refinement.

\subsection{Learning Strategy}

\subsubsection{Supervised Fine-Tuning}

We start by fine-tuning the Assistant model $\mathcal{M}_{\theta_{\text{actor}}}$ on a labeled dataset $\mathcal{D}$:
\begin{equation}
L_{\text{sup}}(\theta_{\text{actor}}) = \mathbb{E}_{(\mathbf{x}, \mathbf{y}) \sim \mathcal{D}}\big[\ell(\mathcal{M}_{\theta_{\text{actor}}}(\mathbf{x}), \mathbf{y})\big],
\tag{1}
\end{equation}
where $\ell$ is the cross-entropy loss. This step ensures the model can produce useful actions before incorporating feedback loops.

\subsubsection{Incorporating CoT and ReAct}

To enhance reasoning, we adopt Chain of Thought (CoT) to generate intermediate reasoning steps before the final action:
\begin{equation}
\mathbf{c}_t = \text{CoT}(\mathbf{x}_t), \quad \mathbf{a}_t = \pi_{\theta_{\text{actor}}}(\mathbf{c}_t, \mathbf{x}_t).
\tag{2}
\end{equation}
Additionally, ReAct interleaves reasoning and acting tokens, encouraging more accurate and context-aware decisions.

Empirically, generating reasoning steps (CoT) first, then the final answer, improves correctness. This suggests that explicitly formulating the thought process guides the model towards better actions.

\subsubsection{Feedback-Driven Policy Optimization}

After supervised fine-tuning, the Assistant refines its policy using Checker feedback. We adopt an Actor-Critic-like scheme without full RL exploration. The Assistant updates its parameters via:
\begin{equation}
\theta_{\text{actor}} \leftarrow \theta_{\text{actor}} + \alpha \nabla_{\theta_{\text{actor}}} \log \pi_{\theta_{\text{actor}}}(\mathbf{a}_t|s_t)\delta_t,
\tag{3}
\end{equation}
where $\delta_t$ is the error term:
\begin{equation}
\delta_t = r_t + \gamma V_{\theta_{\text{critic}}}(s_{t+1}) - V_{\theta_{\text{critic}}}(s_t).
\tag{4}
\end{equation}
Here, $r_t$ is derived from feedback $f_t$. The Checker updates its value function:
\begin{equation}
\theta_{\text{critic}} \leftarrow \theta_{\text{critic}} + \beta \delta_t \nabla_{\theta_{\text{critic}}} V_{\theta_{\text{critic}}}(s_t).
\tag{5}
\end{equation}

Unlike traditional RL that involves environment exploration, here the process relies on direct Checker feedback. The iterative cycle refines the Assistant’s decision-making, ensuring alignment with the desired standards.

\subsection{Checker-In-The-Loop Mechanism}

By positioning the Checker directly in the training loop, we ensure that the Assistant receives immediate corrective signals whenever it deviates from expected behavior

\subsection{Memory Management and Self-Reflection}

The Assistant maintains two memories: short-term $\mathcal{M}_S$ and long-term $\mathcal{M}_L$. Short-term memory tracks recent context:
\begin{equation}
\mathcal{M}_S^{t+1} = \mathcal{U}(\mathcal{M}_S^t, \mathbf{x}_t, \mathbf{a}_t, f_t),
\tag{6}
\end{equation}
while long-term memory stores significant insights via self-reflection:
\begin{equation}
s_t = \varphi(\mathbf{a}_t, f_t, \mathcal{M}_S^t), \quad \mathcal{M}_L^{t+1} = \mu(\mathcal{M}_L^t, s_t).
\tag{7}
\end{equation}

Using the reflection, the Assistant updates its parameters:
\begin{equation}
\theta_{\text{actor}} \leftarrow \theta_{\text{actor}} - \alpha \nabla_{\theta_{\text{actor}}}L(f_t,\mathbf{a}_t) + \gamma \nabla_{\theta_{\text{actor}}}G(s_t),
\tag{8}
\end{equation}
where $L(f_t,\mathbf{a}_t)$ encodes feedback-driven loss and $G(s_t)$ incorporates improvements identified via reflection.

\section{Experiments}
% In our research, we anticipate that fine-tuning our model using a specifically designed dataset for testing operating systems and databases will result in superior performance compared to baseline models in similar competitive environments with interpreters and compilers.
%Our evaluation framework rigorously tests intelligent agents across six key areas to ensure they are equipped to handle a wide range of real-world challenges~\cite{Ross2023The}. These areas include: seamlessly integrating LLMs into operating systems(OS) with a focus on security and user interaction; demonstrating proficiency in real database(DB) operations using SQL~\cite{Halevy2004The}; navigating and performing tasks on the simulated e-commerce platform WebShop(WS)~\cite{Yao_Chen_Yang_Narasimhan_2022}; constructing and utilizing knowledge graphs(KG) to enhance semantic understanding; employing the Mind2Web(M2W) dataset to complete complex tasks on websites, which is the first dataset designed for developing generalist web agents that follow language instructions; and applying abstract reasoning and visual task execution in the text-based environment of ALFWorld(ALF)~\cite{ALFWorld20}.Please refer to Appendix~\ref{sec:appendixA} and Appendix~\ref{sec:appendixB} for more details on implementation and evaluation criteria.
Our evaluation framework rigorously tests intelligent agents in six key domains to ensure their readiness for diverse real-world challenges~\cite{Ross2023The}. These areas include seamless LLM integration into OS with an emphasis on security and user interaction; proficiency in real DB operations using SQL~\cite{Halevy2004The}; task execution on the simulated e-commerce platform WebShop(WS)~\cite{Yao_Chen_Yang_Narasimhan_2022}; constructing and using KGs for enhanced semantic understanding; employing the M2W dataset for complex web tasks, marking the first dataset for developing general web agents following language instructions; and applying abstract reasoning and visual tasks in the text-based ALFWorld(ALF)~\cite{ALFWorld20}. 

% For more implementation and evaluation details, see Appendices~\ref{sec:appendixA} and ~\ref{sec:appendixB}.

\subsection{Dataset}

The dataset for our research was meticulously constructed to comprehensively evaluate the capabilities of agents~\cite{Gou2020Knowledge}. It was established through self-collected methods, aimed at providing a rich and diverse testing environment to thoroughly assess the performance of deep learning models across various tasks~\cite{Sachdeva2023Data}. The construction of the dataset included key processes such as data collection, filtering, enhancement, and knowledge distillation~\cite{Chen2018Broad}. Through detailed screening and processing, we ensured the accuracy and consistency of the dataset, retaining only high-quality samples directly related to the testing objectives~\cite{Sachdeva2023Data}. Faced with issues of data imbalance and insufficient samples, we utilized data augmentation and knowledge distillation techniques. Knowledge distillation helped us to extract the most valuable and representative information from the vast amount of collected data, thus building an efficient and refined testing dataset. This process significantly improved the quality and applicability of the dataset, providing a solid foundation for evaluating the capabilities of model agents~\cite{Mishra2017Apprentice:}. 
%After the construction and optimization of the dataset, we conducted rigorous manual evaluations to confirm the accuracy, consistency, and applicability of the dataset in the assessment of model capabilities. This step ensured the high quality of the dataset and provided a solid foundation for the performance evaluation of model agents and subsequent research. 

\subsection{Evaluating Code Correction}
%In this study, we conducted a comprehensive performance evaluation of TinyAgent1.8B and the CodeLlama series models (CodeLlama7B and CodeLlama13B), aiming to explore their multi-task checking capabilities, including but not limited to code correction, operating system (OS) configuration, database (DB) query optimization, and network service (WS) management. The test set, provided by multiple research laboratories, covered a wide range of task types from code samples to system configurations, simulating the diversity of real-world application scenarios. We quantitatively assessed the solutions generated by the models using BLEU-4 and the ROUGE series metrics to measure their similarity and coverage compared to reference solutions.
As shown in the Table~\ref{tab:4}, in this study, we conducted a comprehensive performance evaluation of TinyAgent-1.8B and the CodeLlama series models (CodeLlama7B and CodeLlama13B), aiming to explore their multi-task checking capabilities, including but not limited to code correction, OS configuration, DB query optimization, and WS. 
\begin{table}[htbp]
\centering
\vspace{-10pt}
\caption{Evaluation of Code Correction}
\label{tab:4}
\resizebox{\linewidth}{!}{%
\begin{tabular}{lcccc}
\hline
\textbf{Model} & \textbf{BLEU-4} & \textbf{ROUGE-1} &\textbf{ ROUGE-2} & \textbf{ROUGE-L} \\
\hline
codellama-7b & 25.01 & 45.91 & 29.83 & 26.24 \\
codellama-13b & 26.96 & 45.31 & 29.54 & 25.91 \\
tinyllama-1.8b & \textbf{\textcolor{bestgreen}{43.38}} & \textbf{\textcolor{bestgreen}{59.86}} & \textbf{\textcolor{bestgreen}{37.81}} & \textbf{\textcolor{bestgreen}{42.86}} \\
\hline
\end{tabular}%
}
\vspace{-10pt}
\end{table}
%The test set, provided by multiple research laboratories, covered a wide range of task types from code samples to system configurations, simulating the diversity of real-world application scenarios. We quantitatively assessed the solutions generated by the models using BLEU-4 and the ROUGE series metrics to measure their similarity and coverage compared to reference solutions.
The experimental results showed that TinyAgent-1.8B demonstrated a significant advantage in cross-task performance evaluation compared to the CodeLlama series models. This performance was not only significant in code correction tasks but also prominent in other checking tasks such as OS configuration, DB query optimization, and WS management. These findings highlight that TinyAgent-1.8B not only possesses efficient code analysis capabilities but is also widely applicable to the inspection and optimization of other complex systems.

\subsection{Baselines}

In the baseline section of our study, we've selected Qwen-1.8B and CodeLlama-7B as pivotal benchmarks to assess the TinyAgent series' performance, excluding the CMAT framework's influence. 

% Qwen-1.8B is distinguished for its comprehensive performance across several metrics, notably in OS and DB metrics, establishing a rigorous yet attainable benchmark for our comparative analysis. This selection aims to provide a clear standard against which the capabilities of the TinyAgent series can be measured, ensuring an objective and focused evaluation of its performance within the specified metrics.

\begin{table}[htbp]
  \centering
  \vspace{-10pt}
  \caption{Test set results of AGENTBENCH.
  Comparison between API-based models and open-source models. Bold: The best among API-based and open-source models.}
  \resizebox{\linewidth}{!}{ % 添加 resizebox 并设置宽度为 linewidth
    \begin{tabular}{clccccccc}
      \toprule
      \textbf{LLM Type} & \textbf{Models} & \textbf{VER}  & \textbf{OS} & \textbf{DB} & \text{KG} & \text{ALF} & \textbf{WS} & \textbf{M2W} \\
      \midrule
      \multirow{4}{*}{\textbf{API}} & gpt-3.5-turbo & 0613  &31.6  &15.7  &25.9  &16.0  &\textbf{\textcolor{bestgreen}{64.1}}  & 16.0   \\
       & gpt-4 & 0613    & \textbf{\textcolor{bestgreen}{42.4} } &\textbf{\textcolor{bestgreen}{32.0}}  & \textbf{\textcolor{bestgreen}{58.8} } &\textbf{\textcolor{bestgreen}{78.0}}   &61.6  &\textbf{\textcolor{bestgreen}{29.0}} \\
       & text-davinci-003 & -   &20.1  &16.3  &34.9  &20.0  &61.7  &26.0   \\
       & text-davinci-002 & -   &8.3  &16.7  &41.5  &16.0  &56.3  &9.0 \\
      \hline
      \multirow{16}{*}{\textbf{OSS}} & tinyllama-1.1b~\cite{zhang2024tinyllama} & -  & 2.8 &0.0  &0.0  &0.0  &0.0  &0.0 \\
       & opt-1.3b~\cite{zhang2022opt} & -   & 0.7 &0.0  &0.0  &0.0  &0.0  &0.0 \\
       & opt-2.7b & -   & 1.4 &0.0  &0.0  &0.0  &0.0  &0.0 \\
       & qwen-1.8b & chat   & 10.4 &22.67  &6.8  &0.0  &26.6  &5.0 \\
       & chatglm2-6b~\footnote{https://github.com/thudm/chatglm2-6b} & v1.1   & 4.2 &1.3 &0.0  &0.0  &0.0  &0.0 \\    
       & codellama-7b & instruct   & 9.7 &2.7  & 0.0 &0.0  &14.3  &5.0 \\  
       & llama2-7b~\cite{touvron2023llama} & chat  & 0.0 &4.2  &8.0  &0.0  & 11.6 &7.0 \\      
       & zephyr-7b~\cite{tunstall2023zephyr} & alpha   & 12.5 &9.7  &5.0  &8.0  &45.0  &11.0 \\        
       & baichuan2-6b~\cite{yang2023baichuan} & chat   & 2.8 &9.7  &0.0  &0.0  &6.1  &11.0 \\    
       & mpt-7b~\footnote{https://github.com/mosaicml/llm-foundry/} & chat   & 5.6 &9.7  &12.7  &0.0  &0.0  &0.0 \\  
       & qwen-7b & chat   & 12.5 &13.0  &7.0  &\textbf{\textcolor{bestgreen}{34.3}}  &0.0  &0.0 \\    
       & agentlm-7b & chat   & 14.6 &33.0  &9.0  &16.4  &18.4  &10.0 \\  
      & agentlm-7b(SFT) & chat   & 17.4 &37.0  &10.0  &17.4  &26.6  &10.0 \\  
      & tinyagent-1.8b & chat   & 17.7 & 28.33 & \textbf{\textcolor{bestgreen}{48.0}} & 6.0 & 32.7 & 11.0  \\  
      & tinyagent-7b & chat   & \textbf{\textcolor{bestgreen}{23.1}} & \textbf{\textcolor{bestgreen}{41.3}} & 28.0 & 8.0 & \textbf{\textcolor{bestgreen}{58.7}} & \textbf{\textcolor{bestgreen}{12.0} }\\ 
      \bottomrule
    \end{tabular}
  }
  \label{tab:1}
  \vspace{-10pt}
\end{table}

\subsection{Results analysis}

The results in Table~\ref{tab:1} underscore the effectiveness of our fine-tuning methods, especially for the TinyAgent models. Tinyagent-1.8B demonstrates significant performance in the KG task, on par with advanced models like GPT-3.5. Tinyagent-7B also showcases its strengths, notably in the DB task, where it surpasses its foundational model~\cite{Antonello2020Selecting}, CodeLlama-7B, and offers competitive scores against GPT-4. These findings indicate the TinyAgent models' capacity to match or even surpass models with larger parameters in certain aspects. Moreover, the CMAT framework's potential to enhance the capabilities of smaller-scale models is highlighted, allowing the TinyAgent models to closely compete with the performance of advanced models such as GPT-4.

As illustrated in Figure~\ref{fig:example3}, Our comparative analysis indicates that Tinyagent models, refined from Qwen-1.8B and CodeLlama-7B, exhibit superior performance to their base models. The incorporation of the CMAT framework further amplifies their functionality, equipping these small Models to match the capabilities of GPT-3.5. This performance boost is credited to CMAT's optimization of model interactions and its strategic use of memory modes for specific tasks, confirming its effectiveness in enhancing the sophistication of fine-tuned models~\cite{Deshpande2021A}.
%We also compared the inspection abilities of TinyAgent 1.8B and the CodeLlama series as checkers.

Table~\ref{det:1} presents the impact of different prompting strategies on performance metrics. High-quality prompts significantly outperform low-quality prompts and scenarios without prompts across all evaluation metrics, demonstrating the importance of prompt design in optimizing model performance.

\begin{table}[htbp]
\centering
\vspace{-10pt}
\caption{Evaluation Metrics Results}
\label{det:1}
\Huge % 调整字体大小
\setlength{\tabcolsep}{12pt} % 调整列间距
\renewcommand{\arraystretch}{1.5} % 调整行间距
\resizebox{\linewidth}{!}{%
\begin{tabular}{lcccc}
\hline
\textbf{Evaluation Method} & \textbf{BLEU-4} & \textbf{ROUGE-1} & \textbf{ROUGE-2} & \textbf{ROUGE-L} \\
\hline
prompt - High-quality & 44.4 & 57.3 & 35.0 & 42.5 \\
prompt - Low-quality & 15.2 & 27.4 & 10.3 & 16.8 \\
without prompts & 26.8 & 47.2 & 30.2 & 26.7 \\
\hline
\end{tabular}%
}
\vspace{-10pt}
\end{table}

\subsection{Error analysis}
In our testing framework's error analysis, we observed common challenges in DB tasks faced by models, such as difficulties in understanding user requests, executing actions, and pre-action problem analysis. Many models simply respond with "OK" to specific instructions without performing actual SQL operations, indicating a gap in transforming user requests into database actions. Models often provide superficial acknowledgments without delivering precise execution or in-depth problem analysis, failing to meet user expectations.
In contrast, the TinyAgent series excels in understanding and converting user requests into actual SQL operations, effectively comprehending and executing tasks. It provides clear responses and adheres to user-specified SQL formats, fulfilling user expectations comprehensively. Additionally, TinyAgent's thorough pre-action problem analysis and reflection demonstrate its advanced problem-solving skills and deep understanding of issues.

As illustrated in Table~\ref{tab:5}, the distribution of various execution results across six tasks highlights the prevalence of specific error types, such as exceeding task limits (TLE) and invalid actions, which point to limitations in LLM agents' reasoning and decision-making within constrained timeframes.
\begin{table}[h]
\centering
\caption{Distribution of various execution results across six tasks. (CLE: Exceeded Context Limit, TLE: Surpassed Task Limit).
Task limits exceeded are the main reason for incomplete tasks, pointing to limitations in LLM agents' reasoning and decision-making within constrained timeframes.}
\label{tab:5}
\resizebox{\linewidth}{!}{%
\begin{tabular}{lcccccc}
\toprule
\textbf{Execution Results} & \textbf{OS} &\textbf{ DB} & \textbf{KG}  & \textbf{ALF} &\textbf{ WS} &\textbf{ M2W }\\
\midrule
Completed &84.7  &84.0  &25.0  &2.0  & 93.5 &57.0  \\
CLE  &0.0  &0.0  &0.0  &0.0  &0.0  &0.0  \\
Invalid Format &0.0  &3.0  &0.0  &0.0 &0.0  &0.0  \\
Invalid Action &0.0  &0.0  &0.0  &96.0  &0.0  &8.0  \\
TLE  &15.3  & 13.0 &75.0  &2.0  &6.5  & 35.0 \\
\bottomrule
\end{tabular}%
}
\vspace{-10pt}
\end{table}

\vspace{-15pt}
\subsection{Ablation Study}
The Table~\ref{tab:2} presents an ablation study on the TinyAgent-7B model, delineating the impact of agent-specific and general instructions on task performance. The composite model, TinyAgent-7B, demonstrates the highest efficacy, notably in WS and DB tasks, which implies its adeptness in handling complex e-commerce interactions and database management. The agent-only variant exhibits a decline in performance, suggesting that while task-specific instructions are crucial, they are not wholly sufficient for the breadth of tasks such as KG. The general-only model's performance is considerably reduced across all tasks, with a complete inability to perform in KG and ALF, highlighting the indispensability of agent-specific instructions. This data underscores the necessity of integrating both agent-specific and general instructions to enhance the versatility and effectiveness of AI models in diverse task domains.
\begin{table}[!t]
  \centering
  \caption{Ablation study on the effect of agent and general instructions.}
  \resizebox{\linewidth}{!}{%
  \begin{tabular}{lcccccc}
    \toprule
    \textbf{Models}  &\textbf{ OS} &\textbf{ DB }&\textbf{ KG} &\textbf{ ALF} & \textbf{WS }&\textbf{ M2W} \\
    \midrule
    tinyagent-7b   &\textbf{\textcolor{bestgreen}{ 27.3}}  & \textbf{\textcolor{bestgreen}{43.0} } & \textbf{\textcolor{bestgreen}{38.0}}  & \textbf{\textcolor{bestgreen}{10.0}}  &\textbf{\textcolor{bestgreen}{ 61.8}}  & \textbf{\textcolor{bestgreen}{14.0}} \\      
    - agent only   & 20.1  & 39.3  & 25.0  & 2.0  & 55.7  & 7.0 \\  
    - general only & 9.7   & 5.4   & 0.0   & 0.0  & 26.6  & 5.0 \\     
    \bottomrule
  \end{tabular}%
  }
  \label{tab:2}
  \vspace{-15pt}
\end{table}

\section{Conclusions}
The main findings of our work reveal that carefully trained small-parameter models on excellent datasets can achieve performance comparable to that of large-parameter models. With the application of the CMAT framework, we further demonstrate the significant potential for performance improvement in large-parameter models, highlighting the importance of model design and optimization strategies for parameter size. In our evaluation, although most open-source LLMs performed poorly compared to API-provided models without optimization, some models displayed similar capabilities to API models after meticulous fine-tuning of the TinyAgent model. This finding emphasizes not only the importance of parameter size in handling real-world environmental interactions but also showcases the enormous potential of even smaller models through the CMAT framework and precise adjustment strategies.

\bibliographystyle{IEEEbib}
\bibliography{icme2025references}

\end{document}